# Hierarchical Label-wise Attention Transformer Model for Explainable ICD Coding

**Leibo Liu[1], Oscar Perez-Concha[1], Anthony Nguyen[2], Vicki Bennett[3], Louisa Jorm[1]**

[1]Centre for Big Data Research in Health, University of New South Wales, Sydney, Australia

[2]The Australian e-Health Research Centre, CSIRO, Brisbane, Queensland, Australia

[3]Metadata, Information Management and Classifications Unit (MIMCU), Australian Institute of Health and Welfare, Canberra, Australian Capital Territory, Australia

*Corresponding Author: Leibo Liu, Centre for Big Data Research in Health, Level 2, AGSM Building (G27) University of New South Wales Sydney, Kensington, New South Wales, 2052 Australia (leibo.liu@student.unsw.edu.au)

# Abstract

International Classification of Diseases (ICD) coding plays an important role in systematically classifying morbidity and mortality data. In this study, we propose a hierarchical label-wise attention Transformer model (HiLAT) for the explainable prediction of ICD codes from clinical documents. HiLAT firstly fine-tunes a pretrained Transformer model to represent the tokens of clinical documents. We subsequently employ a two-level hierarchical label-wise attention mechanism that creates label-specific document representations. These representations are in turn used by a feed-forward neural network to predict whether a specific ICD code is assigned to the input clinical document of interest. We evaluate HiLAT using hospital discharge summaries and their corresponding ICD-9 codes from the MIMIC-III database. To investigate the performance of different types of Transformer models, we develop ClinicalplusXLNet, which conducts continual pretraining from XLNet-Base using all the MIMIC-III clinical notes. The experiment results show that the F1 scores of the HiLAT+ClinicalplusXLNet outperform the previous state-of-the-art models for the top-50 most frequent ICD-9 codes from MIMIC-III. Visualisations of attention weights present a potential explainability tool for checking the face validity of ICD code predictions.

# 1. Introduction

The International Classification of Diseases (ICD) system is maintained by the World Health Organization (WHO) and is widely used to systematically code diseases for different purposes including health statistics, medical reimbursement and billing, decision support and medical and health services research [1, 2]. Currently, trained clinical coders manually perform the ICD coding process. Manual coding is costly, laborious, and error-prone [3-5] and has traditionally suffered from a lack of workforce supply [5]. Thanks to the developments in the fields of natural language processing (NLP) and machine learning techniques, automated ICD coding has been an active research task for over two decades [6, 7] but as yet has not been widely implemented at scale [8].

Automated ICD coding is formulated as a multi-label text classification task that assigns a subset of ICD codes to a given clinical document [9-11]. In recent years, convolutional neural networks (CNNs) and recurrent neural networks (RNNs) have been widely used for automated ICD coding tasks [4, 9, 10, 12-21]. To improve the performance of ICD coding models, an attentive long short-term memory architecture that added a label-wise attention layer upon a hierarchical LSTM network was proposed by Shi et al. [13]. Similarly, a hierarchical attention-bidirectional gated recurrent unit model [12] and a label-wise convolutional attention for multi-label classification [9] were introduced to employ the label-wise attention mechanisms to capture the important sentences or text snippets correlated with each code from the document. Motivated by the success of attention mechanisms with ICD coding, several studies [10, 14, 16, 19] have used different label-wise attention mechanisms with various configurations of CNNs or RNNs and achieved the state-of-the-art (SOTA) results on the publicly accessible dataset Medical Information Mart for Intensive Care (MIMIC-III) [22]. In addition to better performance, attention explainability can be incorporated to increase users' trust in a post-hoc manner by highlighting the text snippets that contribute more to the specific label predictions. Here, label-wise attention mechanisms can be used in models to help explain the reasons why the models assign the subset of codes to the given document by giving different weight scores to different text snippets or words in the document.

The Transformer is an encoder-decoder architecture using stacked multi-head self-attention and position-wise fully connected feed-forward layers [23]. Some studies [24, 25] have embedded the Transformer encoder/decoder in their model architectures to predict ICD codes and achieved better results than the previous CNN- or RNN-based models, but pretrained Transformer-based language models, especially Bidirectional Encoder Representations from Transformers (BERT) [26], have become popular and dominated the SOTA in many NLP tasks [27]. However, pretrained BERT models suffer from a discrepancy problem caused by symbols (such as [MASK]) that are manually added during the pretraining process, but which are not seen in fine-tuning datasets. To resolve this, Yang et al. [28] proposed another type of Transformer-based model, XLNet, a generalized autoregressive pretraining of language understanding. XLNet has been demonstrated to consistently outperform BERT in 20 NLP tasks [28].

Although many studies have attempted to fine-tune the pretrained Transformer-based language models e.g., BERT or its variants in ICD coding tasks [11, 29-33], only a few have investigated the pretrained XLNet and none of these outperformed the SOTA approaches. In addition, BERT imposes a maximum sequence length (512 tokens) to the input data due to the quadratic computational complexity. Therefore, it is a challenge to apply BERT to long documents which have thousands of words. Technically, any length of text sequence can be input into XLNet for finetuning, however the memory usage and computational complexity will quadratically increase as the sequence length grows.

Motivated by the success of label-wise attention mechanisms and Transformer-based models in ICD coding tasks and the robustness of XLNet in many NLP tasks, we propose a Hierarchical Label-Wise Attention Transformer model (HiLAT) for explainable ICD coding. The contributions of the study are:

- Pretrained XLNet models[1]. In the biomedical NLP community, there are some domain-specific BERT models, which were pretrained on biomedical datasets, such as BioBERT [34], ClinicalBERT [35], BlueBERT [36] and PubMedBERT [37]. They have been proven to outperform general-domain pretrained BERT models on many biomedical NLP tasks [34-37]. However, there are only two biomedical domain-specific pretrained XLNet models (ClinicalXLNet [28]), which were pretrained on 1) nursing notes (n=1,077,792) and 2) discharge summaries (n=59,652) from MIMIC-III, respectively. We create two new XLNet models by continual pretraining XLNet from XLNet-Base[2] checkpoint on all the clinical notes of MIMIC-III combined (n=2,083,180) and all the clinical notes excluding the discharge summaries of MIMIC-III (n=2,023,528).
- A HiLAT[3] architecture for explainable ICD coding from discharge summaries. HiLAT takes discharge summaries as input and assigns a subset of ICD codes to each of the discharge summaries. The attention weights produced by HiLAT make predictions explainable through highlighting the clinical text that influenced the assignment of ICD codes.

## 2. Related Work

Research on automated ICD coding can be traced back to the early 1990s [6]. Many rule-based, conventional machine learning and deep learning approaches have been studied by researchers [7, 9, 38, 39]. However, they have recently been outperformed by attentional CNNs/RNNs and Transformer-based approaches [9, 10, 14, 18, 24, 25].

---

[1] The source code of pretraining task and the pretrained models can be downloaded at https://github.com/leiboliu/ClinicalplusXLNet.
[2] XLNet-Base: 12-layer, 768-hidden, 12-heads is available at https://github.com/zihangdai/xlnet.
[3] The source code of HiLAT is available at https://github.com/leiboliu/HiLAT.

## 2.1 Attentional CNNs/RNNs

With the success and effectiveness of attention mechanisms and deep neural networks in NLP tasks, more and more studies have been undertaken to explore the fusion of these two technologies on the ICD coding task and have achieved SOTA results.

Mullenbach et al. [9] proposed an attentional convolutional network (CAML) to automatically assign ICD code sets to discharge summaries using MIMIC-II and MIMIC-III datasets. CAML utilized a label-wise attention layer above a single channel CNN layer to pay attention to the most relevant text within the document that were most relevant for each specific code. A model variant Description Regularized CAML (DR-CAML) was built to improve performance for the rarely observed codes by regularizing the model parameters. Sadoughi et al. [14] improved CAML by using a four channel CNN with maximum pooling across the channels and label-dependent attention layer (MVC-LDA). They further proposed a model variant via regularizing the attention layer (MVC-RLDA) by using ICD code descriptions to enhance MVC-LDA. Li et al. [18] proposed a Multi-Filter Residual Convolutional Neural Network (MultiResCNN) that combined the multi-filter convolutional layer and residual convolution layer with a label-wise attention layer to capture various text patterns with different lengths and enlarge the receptive field for ICD coding.

Baumel et al. [12] presented a Hierarchical Attention bidirectional Gated Recurrent Unit (HA-GRU) network to effectively encode long clinical documents via two levels of GRU layers. A label-wise attention layer was added to the second GRU layer to focus on the relevant sentences for each label. However, HA-GRU could only provide a sentence-level attention for each label. To enhance model explainability, Dong et al. [19] proposed a Hierarchical Label-wise Attention Network (HLAN), which applied a word-level label-wise attention to HA-GRU. Shi et al. [13] used a hierarchical label-wise attention LSTM architecture (AttentiveLSTM) to perform ICD coding. They explored two types of attention mechanism: hard-selection, which selected the maximum attention score for each code, and soft attention, which applied a softmax function to normalize the attention scores. Vu et al. [10] proposed a label attention model (LAAT) and a hierarchical joint learning model (JointLAAT) for ICD coding. The LAAT used a bidirectional LSTM network to produce input feature representations and a label-wise attention layer to learn label-specific vectors of clinical text for each label. The JointLAAT firstly employed a LAAT model to predict normalized codes (the ICD codes' first three characters) and secondly concatenated the projected normalization output with label-specific vectors of another LAAT model to generate the final prediction.

## 2.2 Transformer-based Architecture

Transformer-based architectures, in particular the pretrained Transformer language models, have become popular for a wide range of NLP tasks. Biswas et al. [24] proposed a Transformer-based code-wise attention model (TransICD) that used a Transformer encoder to capture contextual word representations. Similarly, Zhou et al. [25] presented an Interactive Shared representation network with self-Distillation mechanism (ISD) that employed a bidirectional multi-layer Transformer decoder to extract interactive shared representations that were captured from clinical notes via a CNN network. Transfer learning, to fine-tune the pretrained Transformer language models, is the prevalent approach for many NLP downstream tasks [36, 40]. Feucht et al. [30] proposed a description-based label attention classifier (DLAC) to provide explainable ICD coding on discharge summaries. DLAC used pretrained Transformer models to represent the documents and *Word2vec* [41] to encode ICD code descriptions. A label attention classifier was employed to predict ICD codes for a given document. The pretrained model Longformer [42] combined with DLAC (Longformer-DLAC) achieved the best performance in the study. Pascual et al. [32] investigated the fine-tuning of another BERT variant (PubMedBERT) for the ICD coding task, called BERT-ICD. Due to the maximum input sequence length of BERT, they proposed five strategies to split the long text to: first 512 tokens, last 512 tokens, mixed first 256 and last 256 tokens, different fixed-length chunks, and meaningful paragraphs.

## 3. Materials and methods

### 3.1 Datasets and Preprocessing

MIMIC-III is a popular freely accessible database comprising information for 46,520 patients admitted to intensive care units (ICU) at the Beth Israel Deaconess Medical Centre in Boston between 2001 and 2012 [22]. It contains 15 types of unstructured free text clinical notes including hospital discharge summaries, nursing and physician notes, ECG reports, and radiology reports. An average of 35 clinical notes are written by clinical professionals for each hospital stay. Meanwhile medical coders assign an average of 14 diagnosis and procedure ICD-9 codes after patients are discharged from the hospital.

We create three datasets from MIMIC-III: 1) MIMIC-III-CN dataset including all the clinical notes; 2) MIMIC-III-CN-DS dataset comprising all the clinical notes except the discharge summaries; and 3) MIMIC-III-50 dataset, consisting of the discharge summaries which are coded with at least one of the most 50 frequent diagnosis and procedure codes. The first two of these datasets are used for continual pretraining XLNet and the third is used to develop and evaluate proposed models for ICD coding task. The details and descriptive statistics of the three datasets are shown in Table 1.

*Table 1. Statistics of three MIMIC-III datasets for two research tasks. MIMIC-III-CN includes all the clinical notes from the MIMIC-III dataset. MIMIC-III-CN-DS excludes all the discharge summaries from MIMIC-III-CN. Both are used to further pretrain XLNet. MIMIC-III-50 only contains the discharge summaries with at least one of the 50 most frequent diagnosis and procedure codes. MIMIC-III-50 is used for the ICD coding task.*

| Task | Description | Datasets | |
|---|---|---|---|
| | | MIMIC-III-CN | MIMIC-III-CN-DS |
| Pretrain XLNet | Admissions | 58,362 | 58,029 |
| | Number of notes | 2,083,180 | 2,023,528 |
| | Types of notes | Nursing/other (n=822,497) Radiology (n=522,279) Nursing (n=223,556) ECG (n=209,051) Physician (n=141,624) Discharge summary (n=59,652) Echo (n=45,794) Respiratory (n=31,739) Nutrition (n=9,418) General (n=8,301) Rehab Services (n=5,431) Social Work (n=2,670) Case management (n=967) Pharmacy (n=103) Consult (n=98) | Nursing/other (n=822,497) Radiology (n=522,279) Nursing (n=223,556) ECG (n=209,051) Physician (n=141,624)  Echo (n=45,794) Respiratory (n=31,739) Nutrition (n=9,418) General (n=8,301) Rehab Services (n=5,431) Social Work (n=2,670) Case management (n=967) Pharmacy (n=103) Consult (n=98) |
| | Sentences | 39,627,037 | 33,885,717 |
| | Words | 490,193,637 | 410,031,911 |
| | Data size | 2.88GB | 2.40GB |

| Task | Description | Datasets |
|---|---|---|
| | | MIMIC-III-50 |
| ICD Coding | Admissions | 11,368 |
| | Number of notes | 12,808 |
| | Unique ICD codes | 50 |
| | Avg words per admission | 2,188 |
| | Min. words per admission | 89 |
| | Max. words per admission | 10,174 |
| | Avg codes per admission | 15 |
| | Min. codes per admission | 2 |
| | Max. codes per admission | 73 |

We perform preprocessing [43, 44] on MIMIC-III-CN and MIMIC-III-CN-DS datasets. First, we split the text into sentences and convert the words to lowercase. Then the special characters of ==, --, __ and the de-identified brackets are removed from the sentences.

To fairly compare our HiLAT models to others, we use the same split strategies as the ones applied in previous studies [9, 10, 13, 25]. The training, validation and test datasets are extracted according to the pre-prepared hospital admission identifiers[4]. There are 8,066, 1,573, 1,729 admissions in MIMIC-III-50 for training, validation, test, respectively. In addition, there are addendum reports to the discharge summaries for some of the hospital admissions. We combine the addendum reports with discharge summary for each hospital admission. To better understanding the content of discharge summaries, we manually investigate some and find that the discharge summaries have structural sections such as "past medical history", "discharge medications", "allergies", "discharge diagnosis", "attending", and "discharge disposition" as shown in Supplementary Fig. S1. The discharge diagnosis section contains extensive diagnosis information and is located at the end of the whole discharge summary. The discharge summaries are split to 10 chunks sequentially with a fixed sequence length of 510 tokens. The padding tokens are inserted for shorter discharge summaries and longer discharge summaries are truncated. Therefore, there is a risk to miss this important information for ICD coding for the long discharge summaries. We move the sections of "discharge diagnosis", "discharge disposition", and "discharge conditions" to the beginning of the discharge summaries to avoid this risk. The same preprocessing for our pretraining task is applied to discharge summaries. We further remove the words not containing any alphabetic characters.

## 3.2 Pretraining XLNet

We utilize the original implementation[5] of XLNet to pretrain the models. First, the training data is converted to TensorFlow[6] records. Secondly, we further pretrain two XLNet models from the checkpoint of XLNet-Base using two generated datasets: MIMIC-III-CN and MIMIC-III-CN-DS, respectively. XLNet-Base was pretrained on BookCorpus, English Wikipedia, Giga5, ClueWeb 2012-B and Common Crawl by Yang et al. [28]. When the training is finished, the last checkpoint is selected as our final pretrained model. Finally, we use Hugging Face Transformers library[7] to convert the checkpoint to Pytorch model which can be loaded for ICD coding task.

## 3.3 HiLAT

The architecture of our proposed HiLAT model for ICD coding is shown in Fig. 1. Overall, the HiLAT model comprises four layers. Before feeding data into HiLAT, a discharge summary is split into multiple chunks and each chunk has a maximum of 510 tokens. The first layer is a pretrained Transformer language model that creates hidden representations for the tokens of each chunk. The second layer is a token-level label-wise attention layer in which a label-specific chunk representation is produced by applying attention weight scores on the tokens within a chunk. The representations from all the chunks for a specific label are stacked together and fed into the third layer, namely the chunk-level attention layer, to generate label-specific document representations. The last layer is the classifier layer which consists of multiple single feed-forward neural networks (FFNN). Each classifier is used to predict the probability of assigning the specific ICD code to the input discharge summary.

---

[4] The files of the pre-prepared hospital admission identifiers can be downloaded from https://github.com/jamesmullenbach/caml-mimic/tree/master/mimicdata/mimic3.
[5] The source codes are available at https://github.com/zihangdai/xlnet.
[6] The version of TensorFlow we are using is 1.13.1. More information about TensorFlow at https://www.tensorflow.org/.
[7] Transformers library is developed by Hugging Face. More details are available at https://huggingface.co/. Converting TensorFlow Checkpoints follows https://huggingface.co/docs/transformers/converting_tensorflow_models.

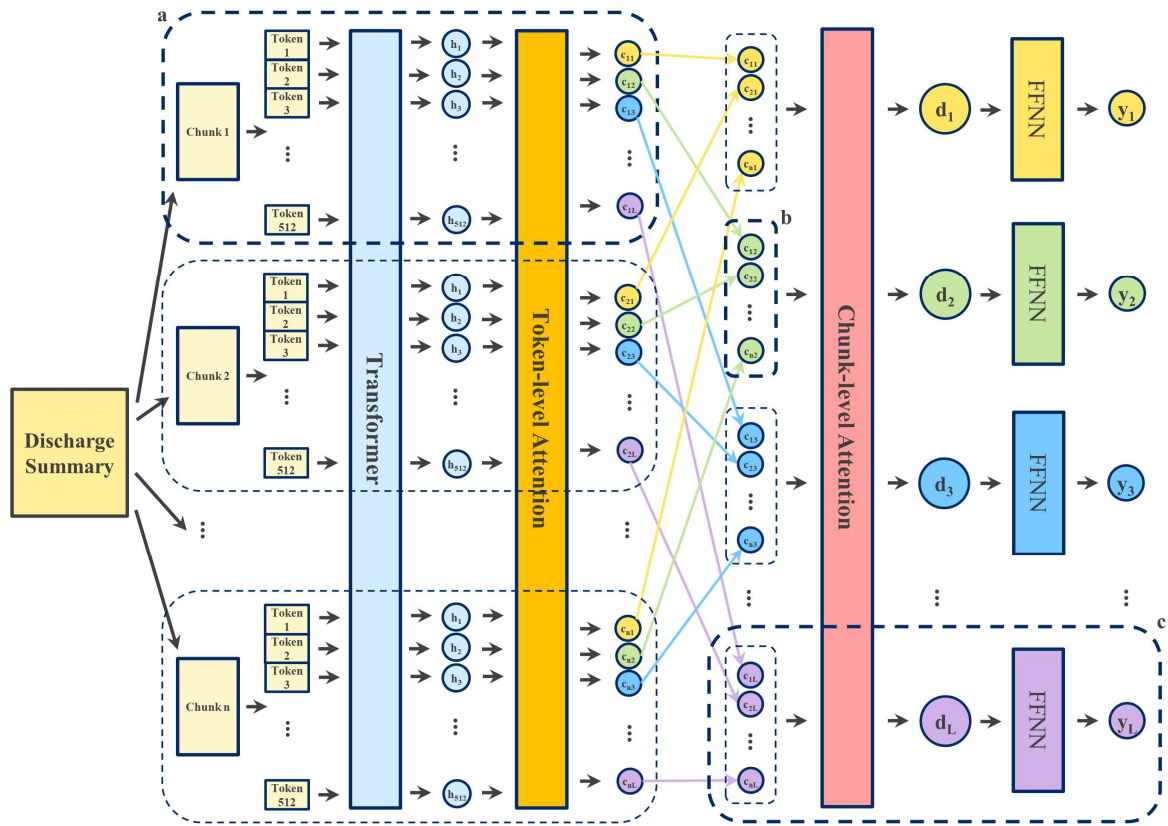

Fig. 1. The architecture of HiLAT. *a,* The text chunk is tokenized using the Transformer tokenizer. The output of the Transformer layer is a matrix $H = [h_1, h_2, h_3...h_{512}]$ which represents the tokens including 510 tokens from text and two special tokens of [CLS] and [SEP]. Given H, the token-level attention layer produces the label-specific chunk representations $C = [c_{11}, c_{12}, c_{13}...c_{1L}]$. L is the number of ICD codes. $C_{12}$ denotes the representation of the chunk 1 for code 2. *b,* The representations for ICD code 2 from n different chunks are combined to a matrix $C = [c_{12}, c_{22}, c_{32}...c_{n2}]$. *c,* Given the label-specific chunk representation matrix, a chunk-level attention layer is employed to produce the document representation for the ICD code L. A binary classifier of FFNN is used to determine whether the ICD code L is related to the input discharge summary or not.

We implement and train HiLAT using Pytorch and Hugging Face Transformers library.

### 3.3.1 Transformer layer

We use the pretrained Transformer models as the Transformer layer and fine-tune all the parameters during the training. The input document $D$ is tokenized by the relevant Transformer tokenizer as a sequence of tokens. We split $D$ to *10* chunks sequentially with a fixed length of 510 tokens for each chunk. The padding and truncating operations are performed for the document shorter and longer than 5,100 tokens, respectively. Each chunk is appended by two special tokens of [CLS] and [SEP] as well. The output of last layer of the pretrained Transformer models is used to represent the tokens $\{t_1, t_2, ..., t_{512}\}$ in chunk $n$. All the token representation vectors are formulated as a matrix $\mathbf{H}_n = [\mathbf{h}_1, \mathbf{h}_2, ..., \mathbf{h}_{512}] \in \mathbb{R}^{d_e \times 512}$, where $d_e$ is the output size of the pretrained Transformer models we are using in HiLAT.

### 3.3.2 Token-level attention layer

The label attention mechanism proposed by Vu et al. [10] is employed in HiLAT. In token-level attention layer, we transform $\mathbf{H}_n$ into $\mathcal{L}$ label-specific vectors representing chunk $n$. $\mathcal{L}$ denotes the size of label set. First, the label-specific attention weights are computed as:

$$\mathbf{Z}_n = \tanh(\mathbf{W}\mathbf{H}_n) \tag{1}$$

$$\mathbf{A}_n = \text{softmax}(\mathbf{U}^T\mathbf{Z}_n) \tag{2}$$

$$\mathbf{C}_n = \mathbf{H}_n\mathbf{A}_n^T \tag{3}$$

$\mathbf{W}$ is a matrix $\in \mathbb{R}^{d_e \times d_e}$. The equation (1) results in a matrix $\mathbf{Z}_n \in \mathbb{R}^{d_e \times 512}$. $\mathbf{U} \in \mathbb{R}^{d_e \times \mathcal{L}}$ is a matrix comprising the $\mathcal{L}$ label representation vectors, which is multiplied with $\mathbf{Z}_n$ to compute the label-specific token-level attention weights $\mathbf{A}_n \in \mathbb{R}^{\mathcal{L} \times 512}$. The $\ell^{th}$ row in $\mathbf{A}_n$ is the attention weights for the 512 tokens in chunk $n$ regarding to the $\ell^{th}$ label in $\mathcal{L}$. To make the summation of the token-level attention weights to equal to 1 for each label, a softmax activation function is applied to $\mathbf{A}_n$. Subsequently, $\mathbf{H}_n$ is multiplied with the transpose of the token-level attention matrix $\mathbf{A}_n$ to generate a matrix $\mathbf{C}_n \in \mathbb{R}^{d_e \times \mathcal{L}}$ for chunk $n$. The $\ell^{th}$ column in $\mathbf{C}_n$ is the label-specific representation of chunk $n$ for the $\ell^{th}$ label in $\mathcal{L}$. We use $\mathbf{c}_{n\ell}$ to denote the label-specific chunk representation for chunk $n$ and label $\ell$.

To generate the input of next layer, each $\ell^{th}$ column is extracted from the matrixes $\mathbf{C}_{1:10}$ to form a new matrix $\mathbf{M}_\ell \in \mathbb{R}^{d_e \times 10}$ ($\ell \in \{1, ..., \mathcal{L}\}$) as:

$$\mathbf{M}_\ell = [\mathbf{c}_{1\ell}, \mathbf{c}_{2\ell}, ..., \mathbf{c}_{10\ell}] \tag{4}$$

### 3.3.3 Chunk-level attention layer

This layer employs the same attention mechanism as the previous one. The matrix $\mathbf{M}_\ell$ is inputted into the chunk-level attention layer to produce the document representation $\mathbf{d}_\ell$ for label $\ell$. The computing equations are as:

$$\mathbf{S}_\ell = \tanh(\mathbf{K}\mathbf{M}_\ell) \quad (5)$$

$$\mathbf{o}_\ell = softmax(\mathbf{v}^T \mathbf{S}_\ell) \quad (6)$$

$$\mathbf{d}_\ell = \mathbf{M}_\ell \mathbf{o}_\ell^T \quad (7)$$

Here, $\mathbf{K} \in \mathbb{R}^{d_e \times d_e}$ is used to multiply with $\mathbf{M}_\ell$ to produce $\mathbf{S}_\ell \in \mathbb{R}^{d_e \times 10}$ using the hyperbolic tangent activation function. The chunk-level attention weight vector $\mathbf{o}_\ell \in \mathbb{R}^{10}$ for label $\ell$ is computed using a randomly initialized vector $\mathbf{v} \in \mathbb{R}^{d_e}$ and the matrix $\mathbf{S}_\ell$ with a softmax function. Finally, the document representation vector $\mathbf{d}_\ell \in \mathbb{R}^{d_e}$ for label $\ell$ is produced using the equation (7).

### 3.3.4 Classifier layer

Given the document representation $\mathbf{d}_\ell$, we use a linear layer as the classifier. The probability $\hat{y}_\ell$ for label $\ell$ is calculated by the classifier with a sigmoid transformation:

$$\hat{y}_\ell = \sigma(\beta_\ell^T d_\ell + b_\ell) \quad (8)$$

where $\beta_\ell \in \mathbb{R}^{d_e}$ is a weight vector and $b_\ell$ is a bias parameter. We use a threshold of 0.5 to predict the binary output for label $\ell$. The training procedure aims to minimize the binary cross entropy loss:

$$L_{BCE} = -\sum_{\ell=1}^{\mathcal{L}} y_\ell \log \hat{y}_\ell + (1 - y_\ell) \log(1 - \hat{y}_\ell) \quad (9)$$

## 3.4 Attention explainability

There are two level attentions in our proposed HiLAT. The token-level attention weights measure the token contribution to the specific label prediction within a chunk. The chunk-level attention weights show the chunk contribution to the specific label prediction within a document. To get the global contribution for each token, we calculate the global token attention weights of chunk $n$ for label $\ell$ as:

$$\mathbf{g}_{n\ell} = \boldsymbol{\alpha}_{n\ell} o_{\ell n} \quad (10)$$

where $\mathbf{g}_{n\ell} \in \mathbb{R}^{512}$ is the global token attention vector for chunk $n$ and label $\ell$. $\boldsymbol{\alpha}_{n\ell} \in \mathbb{R}^{512}$ is the token-level attention weight vector for chunk $n$ and label $\ell$ (the $\ell^{th}$ row of the matrix $\mathbf{A}_n$); $o_{\ell n}$ is the chunk-level attention weight (a scalar) for chunk $n$ and label $\ell$ (the $n^{th}$ element of the chunk attention vector $\mathbf{o}_\ell$).

In Transformer models, the token does not equal to the word because of using different tokenizers such as WordPiece, SentencePiece. The word is split to one or more than one tokens. The global attention vector $\mathbf{g}_{n\ell}$ is based on tokens. To calculate the attention weights of words for explainability visualization, the attention weights of the tokens that belong to the same word are summed together and then normalized as the attention weight for the word. Fig. 2 shows an example of word attention calculation.

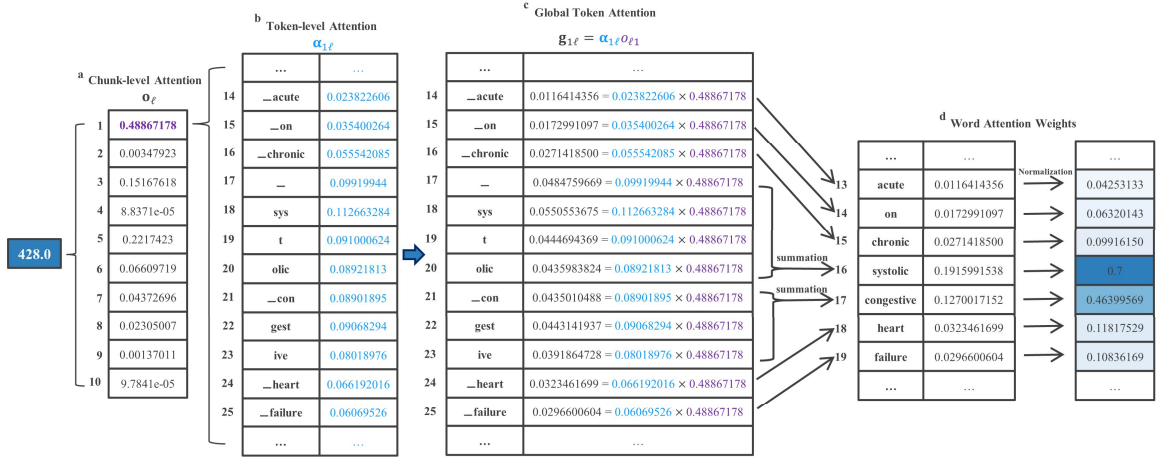

Fig. 2. An example of calculating the word attention weights from chunk and token attention weights. a, The chunk attention weights $\mathbf{o}_\ell$ for ten chunks. b, The snippet of tokens and token-level attention weights $\mathbf{\alpha}_{1\ell}$ from chunk 1. c, The global token attention weights $\mathbf{g}_{1\ell}$ are calculated according to the equation (10). d, The word attention weights are computed by adding the global token attention weights belonging to a word. After that, a normalization is applied to get the final attention weights which are used for explainability visualization.

## 4. Results

### 4.1 Pretrained XLNet models

We train the XLNet models on eight TPU[8] v3 cores with 128GB of total TPU memory for one million steps, using the hyperparameters of batch size of 64, sequence length of 512, learning rate of 4e-4, weight decay of 0.01 and dropout of 0.1. The training processes take about 8.5 days and 8 days to generate the final models on the datasets of MIMIC-III-CN and MIMIC-III-CN-DS, respectively. We name the final pretrained models ClinicalplusXLNet and ClinicalplusXLNet-DS. The performance of the models is evaluated in our ICD coding task.

---

[8] TPU: Tensor Processing Unit. More information is available at https://cloud.google.com/tpu/docs/intro-to-tpu.

## 4.2 ICD coding

### 4.2.1 Metrics and hyperparameters

The previous approaches detailed in related work are selected as our baselines, including CNN-based models (CAML, DR-CAML, MVC-LDA, MVC-RLDA, MultiResCNN), RNN-based models (HLAN, AttentiveLSTM, LAAT, JointLAAT) and Transformer-based models (TransICD, ISD, Longformer-DLAC, BERT-ICD). We evaluate HiLAT using macro- and micro-averaged AUC (area under the ROC curve) and F1 and precision @ 5, 8, 15 to completely compare with baselines.

A manual search of hyperparameters is performed through running eight times of training HiLAT on the MIMIC-III-50 dataset, as shown in Supplementary Table S1. The optimal results are batch size at 16, learning rate at 5e-5, optimizer weight decay at 0.1, training steps at 2,500, and warmup steps at 500. The dropout rate is always set to 0.1. We train the models on eight TPU v3 cores.

### 4.2.2 Model performances

Table 2 shows the evaluation results for HiLAT using MIMIC-III-50. We experiment with six different pretrained Transformer language models in the Transformer layer. We also report the number of parameters and training times.

HiLAT+ClinicalplusXLNet produces the highest scores for all metrics except P@8. Specifically, HiLAT+XLNet-based language models outperform HiLAT+BERT-based language models by a large margin. The domain-specific pretrained language models such as ClinicalplusXLNet improve the macro-AUC by 2%, micro-AUC by 1.2%, macro-F1 by 7.6%, micro-F1 by 3.6%, P@5 by 2.2%, P@8 by 2% and P@15 by 0.9%, compared to the general-domain pretrained language model of XLNet. The language models pretrained on the datasets including MIMIC-III discharge summaries, ClinicalXLNet and ClinicalplusXLNet, perform better than ClinicalplusXLNet-DS pretrained on the dataset excluding discharge summaries. However, the training speeds of HiLAT with XLNet variants are almost four times slower than the ones with BERT variants because of using all possible permutations of the sequences to capture the bidirectional context.

*Table 2. Results on MIMIC-III-50. The bold scores indicate the best results for each metric. \* indicates that the performance difference between the relevant model and the best model on the specific metric is not significant (p > .05) using Approximate Randomization Test [45], which is a non-parametric significance test suitable for NLP tasks.*

| Model | Parameter Number | Training Time (hours) | AUC | | Precision | | Recall | | F1 | | P@k | | |
|---|---|---|---|---|---|---|---|---|---|---|---|---|---|
| | | | Macro | Micro | Macro | Micro | Macro | Micro | Macro | Micro | P@5 | P@8 | P@15 |
| HiLAT + ClinicalBERT | 109,567,538 | 2.27 | 85.6 | 89.7 | 57.8 | 72.9 | 37.7 | 46.8 | 45.6 | 57.0 | 58.9 | 47.5 | 32.3 |
| HiLAT + PubMedBERT | 110,739,506 | 2.43 | 89.0 | 92.3 | **67.5** | **75.8** | 48.0 | 55.7 | 56.1 | 64.2 | 63.9 | 51.6 | 34.2 |
| HiLAT + XLNet | 117,975,602 | 8.31 | 90.7 | 93.8 | 60.3 | 70.4 | 62.6 | 69.4 | 61.4 | 69.9 | 65.9 | 53.4 | 35.3 |
| HiLAT + ClinicalXLNet | 117,975,602 | 8.06 | **92.9** | 94.9* | 65.2 | 69.8 | **71.6** | **76.0** | 68.3 | 72.8 | 67.7* | **55.5** | 36.1* |
| HiLAT + ClinicalplusXLNet-DS | 117,975,602 | 8.85 | 92.4 | 94.7 | 63.5 | 70.4 | 70.6 | 75.1 | 66.9 | 72.7 | 67.9* | 55.1 | 36.0 |
| HiLAT + ClinicalplusXLNet | 117,975,602 | 8.68 | 92.7* | **95.0** | 67.2* | 71.9 | 71.0* | 75.2 | **69.0** | **73.5** | **68.1** | 55.4* | **36.2** |

We compare our proposed HiLAT against 13 baselines as shown in Table 3. HiLAT beats all the baselines on most metrics with a notable improvement. When comparing to the current SOTA model (ISD), HiLAT performs better in three of five evaluation metrics, with an improvement of the micro-AUC, macro-F1 and micro-F1 by 0.1%, 1.1%, and 1.8%, respectively.

*Table 3. Results of comparison with baselines on MIMIC-III-50. The bold scores indicate the best results for each metric. The scores of baseline models are from the corresponding papers.*

| Neural Network | Model | AUC | | F1 | | P@k | | |
|---|---|---|---|---|---|---|---|---|
| | | Macro | Micro | Macro | Micro | P@5 | P@8 | P@15 |
| CNN | CAML [9] | 87.5 | 90.9 | 53.2 | 61.4 | 60.9 | - | - |
| | DR-CAML [9] | 88.4 | 91.6 | 57.6 | 63.3 | 61.8 | - | - |
| | MVC-LDA [14] | - | - | 59.7 | 66.8 | 64.4 | - | - |
| | MVC-RLDA [14] | - | - | 61.5 | 67.4 | 64.1 | - | - |
| | MultiResCNN [18] | 89.9 | 92.8 | 60.6 | 67.0 | 64.1 | - | - |
| RNN | AttentiveLSTM [13] | - | 90.0 | - | 53.2 | - | - | - |
| | HLAN [19] | 88.4 | 91.9 | 57.1 | 64.1 | 62.5 | - | - |
| | LAAT [10] | 92.5 | 94.6 | 66.6 | 71.5 | 67.5 | 54.7 | 35.7 |
| | JointLAAT [10] | 92.5 | 94.6 | 66.1 | 71.6 | 67.1 | 54.6 | 35.7 |
| Transformer | BERT-ICD [32] | 84.5 | 88.7 | - | - | - | - | - |
| | Longformer-DLAC [30] | 87.0 | 91.0 | 52.0 | 62.0 | 61.0 | - | - |
| | TransICD [24] | 89.4 | 92.3 | 56.2 | 64.4 | 61.7 | - | - |
| | ISD [25] | **93.5** | 94.9 | 67.9 | 71.7 | **68.2** | - | - |
| | **HiLAT + ClinicalplusXLNet** | 92.7 | **95.0** | **69.0** | **73.5** | 68.1 | **55.4** | **36.2** |

#### 4.2.3 Ablation Study

To better understand the effectiveness of our proposed approach, we perform an ablation study using our best model on the MIMIC-III-50 dataset. Five important factors are removed or changed to identify the impact on the performance.

1) **Removing non-alphabetic characters**. Instead of removing non-alphabetic characters from datasets, two different settings are evaluated: a) keeping non-alphabetic characters and b) further removing stop words.

2) **Moving important diagnosis information to the beginning and chunking each discharge summary sequentially.** c) Raw text order: We train the model keeping the original text order to evaluate the consequences of possibly missing important diagnosis information at the end of the discharge summaries. d) Meaningful chunking: each chunk contains only the text from specific discharge summary sections. Padding and truncating are applied to each chunk. The sections in each chunk are shown in Supplementary Table S2.

2) **Fine-tuning all the layers of pretrained Transformer model**. e) We only fine-tune the last layer of the pretrained Transformer model rather than fine-tuning all 12 layers.

3) **Randomly initializing attention layer parameters**. The descriptions of ICD codes contain meaningful text. We extract label representations by applying mean pooling operation on the output of the pretrained Transformer model. f) The label representations are used to initialize the parameters of attention layer instead of random initialization.

4) **A single token-level attention layer for all the chunks**. A g) multi-head token-level attention mechanism is implemented and examined. Multi-head means that each chunk has an individual token-level attention layer.

4) **Chunk-level attention**. We evaluate effectiveness of replacing the chunk-level attention layer with h) mean pooling, i) maximum pooling and j) flat concatenation to generate document representations, respectively.

The results of the ablation experiments are shown in Table 4. All of the ablation experiments perform worse than HiLAT+ClinicalplusXLNet. This demonstrates that all of the five factors contribute improvements to our model. Especially, the fine-tuning all Transformer layers and hierarchical label-wise attention mechanism improve the performance by a large margin.

*Table 4. Ablation study on MIMIC-III-50 dataset. The minus value shows how much the performance decreases from the best model.*

| Ablation Study | AUC | | F1 | | P@k | | |
|---|---|---|---|---|---|---|---|
| | Macro | Micro | Macro | Micro | P@5 | P@8 | P@15 |
| **HiLAT + ClinicalplusXLNet** | 92.7 | 95.0 | 69.0 | 73.5 | 68.1 | 55.4 | 36.2 |
| a) keeping non-alphabetic characters | 0.0 | -0.1 | -0.5 | -0.2 | -0.2 | 0.0 | 0.0 |
| b) further removing stop words | -0.4 | -0.3 | -1.9 | -1.3 | -0.4 | -0.6 | -0.3 |
| c) raw text order | -0.3 | -0.1 | -0.2 | -0.1 | -0.1 | -0.1 | 0.0 |
| d) meaningful chunking | -1.6 | -1.2 | -4.0 | -2.9 | -1.9 | -1.8 | -0.8 |
| e) only fine-tuning the last layer of Transformer model | -1.9 | -1.4 | -7.0 | -3.6 | -2.3 | -2.2 | -1.2 |
| f) label embedding | -0.2 | -0.2 | -0.5 | -0.8 | -0.8 | -0.3 | -0.1 |
| g) multi-head token-level attention | -1.3 | -0.8 | -3.5 | -2.2 | -1.3 | -1.0 | -0.7 |
| h) mean pooling for document representations | -2.8 | -2.0 | -12.7 | -8.9 | -4.4 | -3.4 | -1.3 |
| i) maximum pooling for document representations | -0.2 | -0.2 | -1.7 | -0.7 | -0.4 | -0.3 | -0.1 |
| j) flat concatenation for document representations | -1.2 | -1.1 | -5.0 | -3.4 | -1.8 | -1.5 | -0.9 |

## 4.3 Model Explainability

The attention weights of the hierarchical label-wise attention mechanism of HiLAT can be used to tag the words in the chunks as relevant to each specific code. Fig. 3 shows examples of model explainability. For the ICD-9 diagnosis codes "276.2 Acidosis" and "428.0 Congestive heart failure, unspecified", our model successfully tags the keywords "metabolic acidosis … increased anion gap metabolic acidosis" and "acute on chronic systolic congestive heart failure", respectively. The keywords "cardiac catherization on and second cardiac catherization on with bms" and "cardiac catherization where two stents" are highlighted for the ICD-9 procedure code of "88.56 Coronary arteriography using two catheters".

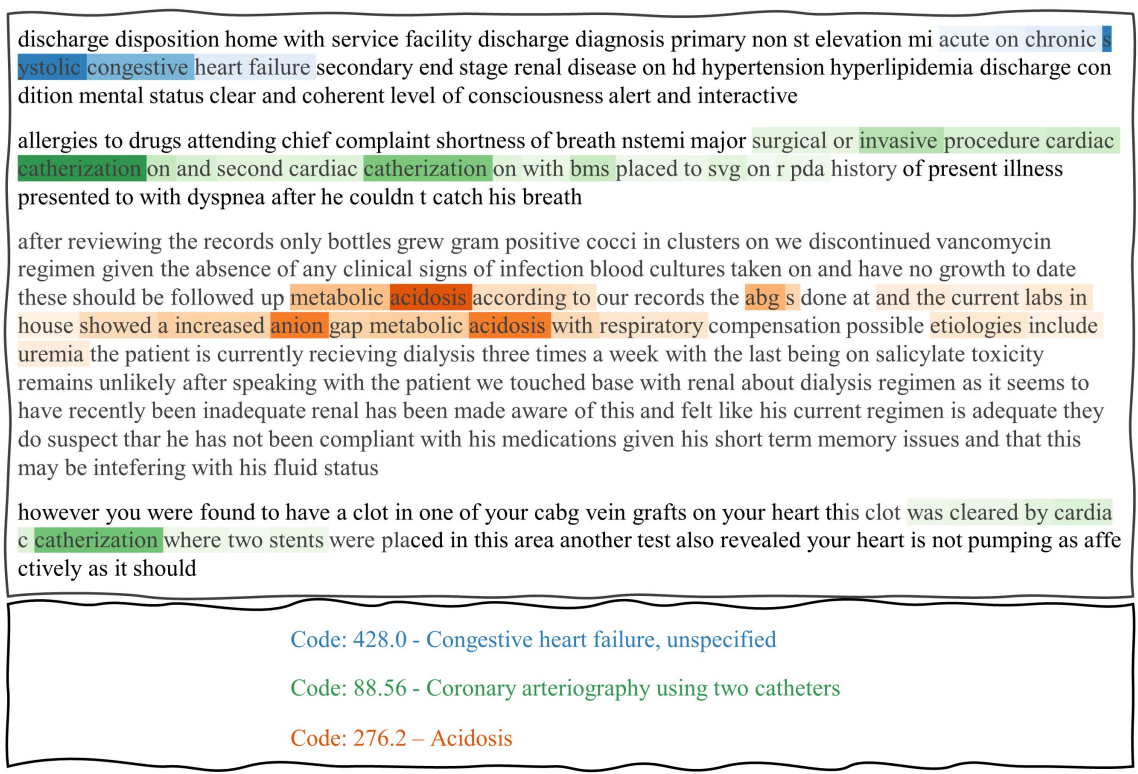

*Fig. 3. Attention visualization examples. Attention visualizations for the code 276.2, 428.0 and 88.56 in one discharge summary are displayed using different colours. The darker colour indicates the higher attention score.*

We also investigate some codes predicted by our model that were not coded by clinical coders, as shown in the first four examples of Fig. 4. The first two examples are from one hospital admission. The patient with chronic lymphocytic leukemia was admitted to ICU due to fatigue and weight loss.

Our model assigns the ICD-9 diagnosis code "518.81 Acute respiratory failure" to the hospital admission according to the keyword "dyspnea". However, the discharge summary does not specifically mention respiratory failure, and this would not be coded on the basis of the symptom of dyspnea alone. The ICD-9 diagnosis code "995.92 Severe sepsis" is predicted according to the highlighted keywords of "possible septic physiology". The meaning of the keyword "possible" is not learnt by our model correctly. The next two examples are from a patient who had shortness of breath and was intubated for respiratory distress. Our model correctly identifies that the patient was intubated during the ICU stay. However, coding standards specify that the ICD-9 procedure code "96.04 Insertion of endotracheal tube" is recorded only when intubation occurs without ventilation, demonstrating challenges for learning when a code should not be recorded on the basis of somewhat arbitrary coding rules. The ICD-9 procedure code "96.71 Continuous invasive mechanical ventilation for less than 96 consecutive hours" is predicted by our model, while the clinical coder assigned "96.72 Continuous invasive mechanical ventilation for 96 consecutive hours or more" to the hospital stay as the patient was extubated after 7 days. Models that incorporate temporal understanding in free text are required to correctly learn such information. The procedure code "96.6 Enteral infusion of concentrated nutritional substances" in the last example of Fig. 4 is predicted correctly by our model, but the highlighted text snippets do not provide meaningful explainability.

| Predicted Code | Highlighted Keywords |
|---|---|
| Code: 518.81<br>Acute respiratory failure | he was extubated on and did well in terms of respiration since then his course in icu complicated by sinus bradycardia thought to be related to amiodarone loading and hypothermia now improved to 60s since being on the floor active issues have included gross volume overload ascites and lll collapse dyspnea while in the unit dyspnea thought to be possibly multifactorial of pneumonia versus exacerbation of asthma component of copd most likely fluid overolad from treating hypotension below hydrocortisone na succ mg iv q6h for asthma or allergic reaction to cefepime or adrenal insufficiency while on the floor the patient had repeat ct chest which showed evidence of lll collapse he was evaluated by pulmonary who thought that a bronchoscopy was not necessarily indicated and instead recommended symptomatic meausures |
| Code: 995.92<br>Severe sepsis | patient recovered his temperature normalized while on the bmt floor the patient remained afebrile hypotension cmv adrenalitis while in the the patient was found to be hypotensive and he was maintained on pressors to keep map he was started on broad spectrum antibiotics and an infectious work up was pursued the patient was also started on hydrocortisone and fludrocortisone per endocrine rec as his blood pressures improved he was weaned off pressors … acute on chronic kidney injury cr improving acute component likely caused by hypoperfusion in the setting of possible septic physiology but could also be related to medications the patient s creat normalized |
| Code: 96.04:<br>Insertion of endotracheal tube | major surgical or invasive procedure intubation bal history of present illness<br>he became obtunded and was intubated he continued to decline cxr showed worsening b l infiltrates<br>presented to an osh for sob and confusion intubated for respiratory distress now extubated ams likely multifactorial with component s of baseline<br>he had been intubated at the osh prior to transfer and we attempted weaning him off the vent over the course of his micu stay |
| Code: 96.71:<br>Continuous invasive mechanical ventilation for less than 96 consecutive hours | major surgical or invasive procedure intubation bal<br>he had been intubated at the osh prior to transfer and we attempted weaning him off the vent over the course of his micu stay and w ere able to wean him to pressure support the patient self extubated three days prior to transfer to the floor |
| Code: 96.6 - Enteral infusion of concentrated nutritional substances | her spinal metastases compression fractures and pain post operatively she was kept in the icu she was initially unable to be extubated and she was found to have a pneumonia and was treated with broad spectrum antibiotics for four days after four days she had a subsequent bal which was negative for any organisms her antibiotics were stopped and she was successfully extubated she was then transferred to the floor where she did well she was discharged to rehab in stable condition medications on admission cyclobenzaprine mg tid dexamethasone mg daily taper last dose today |

*Fig. 4: Examples of codes predicted by our model that were not coded by clinical coders (the first four examples) and code predicted by our model correctly (last example) of that the highlighted text does not provide evidence for the prediction. The darker colour indicates the higher attention score.*

## 5. Discussion

In this study, we propose a hierarchical label-wise attention Transformer model (HiLAT) to automatically predict ICD codes from discharge summaries. In addition, we pretrain two XLNet language models: ClinicalplusXLNet and ClinicalplusXLNet-DS using the clinical notes of MIMIC-III. The experiment results show that Our HiLAT+ClinicalplusXLNet model outperforms all the baseline models in macro- and micro-F1 scores by a large margin for predicting ICD-9 codes on the MIMIC-III-50 dataset, comprising the discharge summaries that are coded with at least one of the most 50 frequent diagnosis and procedure codes. The ablation study demonstrates that all the components of our proposed approach contribute improvements to the model performance.

Explainable ICD coding will build users' trust in automated coding models and help to remove obstacles to deploying automated coding systems in practice [46, 47]. We demonstrate that visualisations highlight the keywords for assigning specific ICD codes according to the hierarchical attention weights, which present a potential explainability tool for checking the face validity of ICD code predictions. Our study shows that stop words do not need to be removed during the preprocessing stage for fine-tuning Transformer language models. Only words without alphabetic characters (e.g., various numbers in the discharge summaries) are removed in the training data to keep relatively complete text for ICD coding. Additionally, the pretrained Transformer language models are trained on large datasets with limited preprocessing so that they have the capability to represent tokens in different contexts.

The main challenge in fine-tuning Transformers on long text datasets, is the limitation of the sequence length. The pretrained BERTs and XLNets are all based on a fixed sequence length of 512 tokens. To address this limitation in our proposed HiLAT, the long text is split into small chunks that are fed into model sequentially and considered as independent contexts. Several Transformer variants such as Longformer [42] and BigBird [48] have proposed different attention mechanisms to handle sequences of length up to 4,096 tokens. Long sequence will allow Transformer models to take the entire discharge summary as input without splitting it into small chunks first. Applying these Transformer variants in HiLAT will be part of our future work.

Training time is another challenge in fine-tuning Transformers. There are about 110 million parameters in the Transformer models. As a result, the training time is much longer than for other types of neural networks. In our experiments, the average training time is about 8.5 hours for XLNet and about 2.5 hours for BERT on the MIMIC-III-50 dataset on eight TPU v3 cores. Therefore, fine-tuning Transformer-based models on long documents efficiently will be an important area for future research, given that the clinical documents are normally quite long (the average length of discharge summaries in our MIMIC-III-50 dataset was 2,188 words).

We tried to train HiLAT on all discharge summaries in MIMIC-III, which contains an extremely large number of ICD-9 codes (n=8,929). We encountered a memory leaking problem after training for about 10 hours due to the model network graph in the loss backward step being too large for the physical memory storage. In an empirical comparison study [49], Yogarajan et al compared fine-tuning three pretrained BERT models with traditional neural networks such as CAML for the ICD coding task, using five different label sizes on two datasets. They found that the Transformer models could not outperform traditional neural networks when the label size was greater than 300. Exploration of the utility for the ICD coding task of Transformer variants, which have been studied for extreme multi-label classification in particular X-Transformer [50] and XR-Transformer [51], could be a fruitful avenue for further research.

Although HiLAT provides label-wise explainability visualizations using the hierarchical attention weights, there is a need to evaluate the quality of explanations objectively. This will help contribute knowledge towards the debate about the usefulness of attention mechanisms for explainability [52, 53]. This will be a crucial research direction of our future work. We plan to explore the use of the mutual information metric [54] to compare the keywords assigned high attention weights by HiLAT with the ground truth of ICD codes, including the ICD code descriptions and the other ICD code mapping terms such as Systemized Nomenclature of Medicine Clinical Terms (SNOMED CT) and Unified Medical Language System (UMLS).

## 6. Conclusion

We present an approach for automated ICD coding from discharge summaries: the HiLAT model with a hierarchical label-wise attention mechanism plus a pretrained Transformer language model, ClinicalplusXLNet. HiLAT extracts the label-wise text representations from discharge summaries to map the corresponding ICD codes. We also demonstrate use of the label-wise attention weights produced by HiLAT to highlight the relevant keywords contributing to its specific ICD code predictions. HiLAT can potentially be applied to different types of multi-label text classification tasks to achieve SOTA results, especially in the clinical health domain with the help of our pretrained language model ClinicalplusXLNet. HiLAT can be deployed to augment and streamline current manual processes for clinical coding, noting that our results relate to performance for predicting the 50 most frequent ICD codes only. Other applications with immediate promise include automated identification from clinical notes of patients with specific conditions who are eligible for recruitment into clinical trials, and of specific clinical endpoints (e.g., major cardiovascular events) for clinical trials and real-world evidence studies.

**CRediT authorship contribution statement**

**Leibo Liu:** Conceptualization, Methodology, Software, Model Building, Evaluation, Writing - original draft. **Oscar Perez-Concha:** Conceptualization, Methodology, Supervision, Writing - review & editing. **Anthony Nguyen:** Methodology, Supervision, Writing - review & editing. **Vicki Bennett:** Methodology, Supervision, Writing - review & editing. **Louisa Jorm:** Conceptualization, Methodology, Supervision, Writing - review & editing.

**Declaration of Competing Interest**

The authors declare that they have no known competing financial interests or personal relationships that could have appeared to influence the work reported in this paper.

**Funding**

This study was supported by the Australian government and the Commonwealth Industrial and Scientific Research Organisation (CSIRO) through Australian Government Research Training Program scholarship and CSIRO top up scholarship.

## Acknowledgements

We thank TPU Research Cloud (TRC) program (https://sites.research.google/trc/about/) for providing free cloud TPUs to support and accelerate our study. This study was supported by the Australian government and the Commonwealth Industrial and Scientific Research Organisation (CSIRO) through Australian Government Research Training Program scholarship and CSIRO top up scholarship.